\documentclass[runningheads]{llncs}
\usepackage[T1]{fontenc}

\usepackage{cite}
\usepackage{graphicx}

\usepackage{xcolor}
\begin{document}
%
\title{Data Poisoning: An Overlooked Threat to Power Grid Resilience}
%
%
\author{Nora Agah\inst{1}\orcidID{0000-0003-0271-1204} \and
Javad Mohammadi\inst{1}\orcidID{0000-0003-0425-5302}
\and
Alex Aved\inst{2} \orcidID{0000-0002-7015-7359}
\and
David Ferris \inst{2} \orcidID{0000-0002-4250-3023}
\and
Erika Ardiles Cruz \inst{2} \orcidID{0000-0003-3092-4696}
\and
Philip Morrone\inst{2} \orcidID{0009-0008-2004-2458}
}
\authorrunning{N. Agah et al.}
%
\institute{$^{1}$The University of Texas at Austin, Austin TX 78712, USA \\
\email\{norakagah, javadm\}@utexas.edu\\
$^{2}$The U.S. Air Force Research Laboratory, Rome, NY 13441, USA \\ 
\email\{alexander.aved, david.ferris.3, erika.ardiles-cruz, philip.morrone.6\}@us.af.mil\\
}
\maketitle              
\begin{abstract}
As the complexities of Dynamic Data Driven Applications Systems increase, preserving their resilience becomes more challenging. For instance, maintaining power grid resilience is becoming increasingly complicated due to the growing number of stochastic variables (such as renewable outputs) and extreme weather events that add uncertainty to the grid. Current optimization methods have struggled to accommodate this rise in complexity. This has fueled the growing interest in data-driven methods used to operate the grid, leading to more vulnerability to cyberattacks. One such disruption that is commonly discussed is the adversarial disruption, where the intruder attempts to add a small perturbation to input data in order to “manipulate” the system operation. During the last few years, work on adversarial training and disruptions on the power system has gained popularity. In this paper, we will first review these applications, specifically on the most common types of adversarial disruptions: evasion and poisoning disruptions. Through this review, we highlight the gap between poisoning and evasion research when applied to the power grid. This is due to the underlying assumption that model training is secure, leading to evasion disruptions being the primary type of studied disruption. Finally, we will examine the impacts of data poisoning interventions and showcase how they can endanger power grid resilience.

\end{abstract}
\section{Introduction}
\vspace{-.2cm}
Data-driven information processing methods are being used with increasing frequency in different Dynamic Data Driven Applications Systems (DDDAS), such as the power grid. Preserving the power grid's resiliency becomes more challenging as extreme weather events and renewable energy usage increase. While data-driven methods may help operators manage the growing variability in the grid, they also bring new potential vulnerabilities. One such vulnerability is the adversarial disruption \cite{szegedy2014intriguing}. Adversarial disruptions involve the adversary using some knowledge or prediction of the victim's data, and adding a small perturbation to that data that deceives the optimization model. Adversarial disruptions have drawn attention across multiple domains \cite{Yao2022Review, Yuan_2021_ICCV,Gnanasambandam_2021_ICCV,li2020bertattack,abbasi2023brainwash,morris2020textattack,li2021contextualized}. Two of the most popularly explored types of adversarial disruptions are poisoning and evasion, which are distinguished mainly based on when the disruption occurs in the process of training and testing the optimization methods \cite{bai2021recent}.

Understanding and addressing adversarial disruptions is important to the deployment of machine learning (ML) to DDDAS such as the power system. Application to the power grid requires fast and frequent updates to keep up with the dynamic nature of the power system, especially as more renewable energy is added to the grid and extreme weather events become more common. These frequent changes require the use of data-driven methods, leading to more opportunities for disruptions. 
To build reliable Artificial Intelligence (AI) for power systems, it is critical to understand adversarial disruptions on the power grid. 
%
Applications of adversarial disruptions to the power system include disrupting supply and demand prediction, grid operation, and disruption detection algorithms. We discuss these applications for both evasion and poisoning, as well as the shortcomings of poisoning literature compared to evasion literature, which previously has been overlooked. 


\section{Problem Formulation}
\vspace{-.2cm}

Adversarial disruptions for evasion and poisoning follow a similar structure, first described by \cite{szegedy2014intriguing}. The intruder either has knowledge of the original data set, which is called a white box attack, or has a method to estimate the data set, which is called a black box attack. Using this knowledge of the dataset, the intruder completes iterative training to fine-tune a small perturbation, $\delta$,  that can be added to the data resulting in the maximal loss of the optimizer while avoiding being detected by bad data detection flags. Neural networks are extremely vulnerable to this type of disruption \cite{szegedy2014intriguing}. The formulation, which is modified from \cite{bai2021recent}, is shown below:

\begin{center}
$max_\delta \; L(\theta, x + \delta, y)$
\end{center}

Where $L$ is the loss function, $\delta$ is the perturbation, $\theta$ is the model parameters, $x$ is the model input, and $y$ is the model output. Poisoning disruptions are when this small perturbation is added to the training data, leading to a known bias that can be exploited. Evasion disruptions are when this perturbation is added at test/run time, taking advantage of the fact that the model is not perfectly fit to the data. This difference is shown in Figure 1.

\begin{figure}[htb!]
\centering
\vspace{-.1cm}
\includegraphics[scale = 0.5]{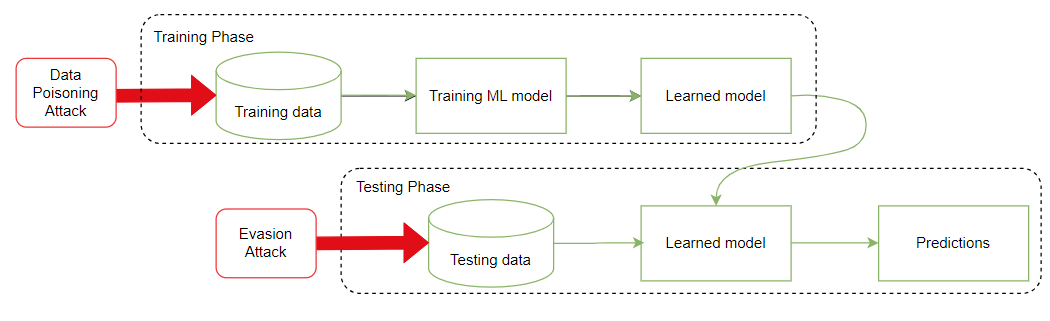}
\caption{The main difference between poisoning and evasion lies in where in the process the disruption occurs. This image is adapted from \cite{Koffas_2022}.}
\label{DifferenceImage}
\end{figure}

There are multiple ways to defend against adversarial disruptions, the most effective being adversarial training \cite{bai2021recent}. This includes adding some adversarial perturbed data in every training iteration, and training to minimize the loss done by the adversary. This minimax problem, formulated in \cite{bai2021recent} is as follows:

\begin{center}
$min_\theta \; max_\delta \; L(\theta, x + \delta, y) $
\end{center}

\section{Evasion Literature}
\vspace{-.2cm}
There are multiple areas of the power grid in which data-driven methods are being used, and a wide range of research is exploring these areas. There are three main categories in which evasion disruptions fall into: supply and demand, grid operation, and security, as shown in Figure 2. These applications are discussed in this section.

\begin{figure}[htb!]
\centering
\includegraphics[scale = .85]{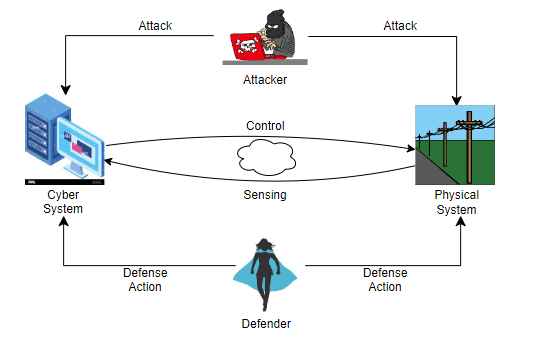}
\vspace{-.5cm}
\caption{There are many components involved in running a power system safely, especially with more cybersecurity becoming necessary. This image is adapted from \cite{Human2014Rege}.}
\label{DifferenceImage}
\vspace{-.7cm}
\end{figure}

\vspace{-.2cm}
\subsection{Targeting Electricity Consumption and Production}
\vspace{-.2cm}
One application of evasion disruptions is modifying demand response predictions. In \cite{chen2019exploiting}, they sought to show the susceptibility to the disruption on load forecasting algorithms. When load forecasting, the forecasting algorithm pulls data from a weather API. This gives the intruder an easy access point for perturbation injection. False weather data of even a degree difference in temperature can cause large overestimation or underestimation of the load. Overestimation may cause generators to be turned on when not needed or encourage tapping into storage, wasting energy and money. Underestimation can lead to load shedding as a result of not preparing to have the correct amount of energy. There is a need for the intruder to inject only a small disturbance to prevent being caught by the bad data detection flags. The injection of false data is executed in two ways. The first is having access to previous load and weather data, as an intruder would in a white box attack, and using it as an estimator to inject a small disturbance. The second is querying the load forecasting algorithm to characterize the data, as an intruder would in a black box attack, in order to add a small disturbance which was enough to create a large variation in load estimation. Zeng et al. propose the robust adversarial multi-agent reinforcement learning framework for demand response (RAMARL-DR) in \cite{Zeng2022multi}. They first show that the multi-agent reinforcement learning framework for a demand response system (MARL-DR) is vulnerable to evasion disruptions, and then use adversarial training to make the controller robust against them. In the multi-agent approach, for disruption scenarios, one of the agents becomes an intruder with the objective of maximizing loss within the system, while the other agents attempt to minimize loss. The adversarial training of RAMARL-DR is formulated as a Markov game, with a series of different agents serving as the intruder, which is shown to make the method robust. DR and load forecasting are critical in balancing supply-demand, and this work in evasion disruptions helps ensure these predictions are useful.

\vspace{-.2cm}
\subsection{Adversarial Interventions in Grid Operation}
\vspace{-.2cm}
A major target for adversarial intruders is grid operation. Information processing protocols are at the core of the power grid's daily operation.  Pan et al. work to improve Reinforcement Learning (RL) control agents for a renewable system that are in charge of decisions to minimize the operation costs of the power grid by changing connections between power lines or generator set points \cite{pan2021improving}. They use both black and white box adversarial training to maximize operation costs. They trained using the problem setup/environment of the Learn to Run a Power Network (L2RPN) competition and trained the adversary on the competition winners. The adversary learns by detaching lines and learning which detachments cause the most damage. They implemented the training on the winning agent, which led to an improvement in its resilience.

Another example is for N-k security-constrained optimal power flow (OPF). This means modeling for k equipment failures, such as generator or line failures, as discussed in \cite{donti2021adversarially}. Most grid operators require the grid to be N-1 secure, but a higher level of security is necessary due to an increase in outages. In \cite{donti2021adversarially}, the defender is trying to schedule power generation to minimize cost, while the intruder is trying to find the equipment failure that will maximize the cost of generation and instability. Ensuring N-k security is a combinatorial problem to account for every combination of failures. The authors propose using projected gradient descent with implicit functions that include the modeling of the physics of the power grid in order to decrease solution convergence time. They test up to N-3 security, which takes about 21 minutes on a standard laptop, and results in 3-4 times fewer feasibility violations than a baseline OPF approach. This example also ties into security, which is discussed in the next section.

\vspace{-.2cm}
\subsection{Misleading the Detection Algorithms}
\vspace{-.2cm}
Adversarial disruptions can also be used in applications for grid security. One such application is misleading disruption detection algorithms. In \cite{Huang2023Mitigation}, Lweet et al. discuss misleading the disruption detection models by inserting a small disturbance in the data used to test them. The authors propose learning a causal relationship for disruption detection to avoid adversarial training due to its high computational needs. Badr et al. in \cite{Badr2023Theft} approach security from a different standpoint: that of electricity theft detection. The machine learning detectors used to flag electricity theft are shown to be tricked using a Generative Adversarial Network that can generate fake low electricity usage readings which allow electricity theft to go unnoticed. Authors in \cite{El-Toukhy2024RL} consider the same electricity reading, with this paper using  Double Deep Q-Network and the Fast Gradient Sign Method to generate adversarial samples. Another similar security application is discussed in \cite{Sayghe2020State}, where the authors show how the False Data Injection Attack Detectors can be tricked using evasion disruptions. They use Limited-memory Broyden-Fletcher-Goldfarb-Shanno (L-BFGS) and the Jacobian-based Saliency Map Attack (JSMA) in order to generate adversarial examples. In \cite{CNN2022Tian}, the authors trick the deep neural network used to monitor power quality and identify power quality issues using an adversarial signal on the model during test time. The authors propose adversarial training to make the model more robust. The literature on evasion disruptions towards the power system continues to grow,
allowing more insight into the extent of the weaknesses in data-driven methods. 

\section{Poisoning Literature and Shortcomings}
\vspace{-.2cm}
\subsection{Poisoning Production and Consumption Predictions}
Similar to evasion work, there has been some research on poisoning disruptions applied to load forecasting. Liang et al. in \cite{Yi2019Load} modify training load data, temperature data, and both load and temperature data to show that load forecasting is vulnerable to all three, even with disruption detectors in place. In \cite{Qureshi2022Poisoning}, the authors disrupt the Long-Short Term Memory(LSTM) model used for federated learning for load forecasting. Federated learning, where multiple agents work to train a model while keeping their own data, works well for load forecasting, as customers are hesitant to share their load data for privacy reasons. This increases the security of the model because the impact of a single agent being contaminated is decreased, but it also leaves more disruption surfaces. The disruption is carried out by assuming some of the agents have been disrupted and the updated weights being sent to the centralized server during training have been poisoned. The authors then propose a spectral clustering algorithm to detect poisoning disruptions. 

\vspace{-.2cm}
\subsection{Poisoning Grid Operation Tools}
\vspace{-.2cm}
There has been a bit of research using poisoning for grid operation, such as in the case of \cite{Gunn2022Pricing}, where the RL agent is in charge of pricing energy costs to maximize profit. The poisoning method Gunn et al. chose was to perturb the trajectory in order to reverse the direction of the estimated gradient. Poisoning disruptions can stagnate the agent's learning, cause the agent to lose money, or increase peak demand. 
There are also papers dedicated to manipulating a model and improving the efficiency of poisoning disruptions, as they are computationally expensive. One such paper is \cite{Zhu2023Edge}, which moves data poisoning online to decrease runtime. The authors focus on poisoning the environmental data used to predict electricity output from a power plant. An incorrect estimate will make it impossible for an operator to correctly balance supply and demand. 

\vspace{-.25cm}
\subsection{Poisoning Power System Security Protocols}
\vspace{-.2cm}
There has been a fair bit of exploration of data poisoning applied to smart grid security, with the focus mostly on sensor/measurement data. In \cite{Islam2022Anomoly}, the authors discuss using data poisoning to disrupt the anomaly-based disruption detection system used in smart-metering. The authors compare data poisoning on different parts of the data, including the smart metering data and the residual data which keeps track of the difference between the safe margin of the data and the sample itself, with both showing a negative impact on the detector. Authors propose modifications to the detector in order to make it more robust to these disruptions. This susceptibility to poisoning of metering data is echoed in \cite{Takiddin2021Electricity}, as the authors show that data poisoning can trick the electricity theft detectors used in smart neighborhoods. In \cite{Hong2023Data}, the authors use data poisoning to manipulate the bad data detection algorithm, similar to the papers in the evasion space. Finally, in \cite{Kamal2021Synch}, the authors discuss the vulnerability of the event classification using distribution-level phasor measurement units (PMU), which measure frequency and help operators determine whether there has been an event. 

Through this work, there did appear to be some overlap between false data injection attacks and data poisoning, but data poisoning appeared to be a term more specific to machine learning and involved the perturbation methods, while false data injection did not. As discussed in these papers, data poisoning has large consequences for data-driven methods in the power system. However, poisoning disruptions have many more applications and variations left to explore compared with evasion disruptions. Additionally, fewer of the poisoning papers versus the evasion papers include new solutions to increase robustness against disruptions. There are multiple different avenues and methods to poison a single model, and it is important to be aware of these, so there can be efforts to make models secure.

\section{Conclusion and Future works} 
\vspace{-.25cm}
%
Through this literature review, we have highlighted the range of power system applications which adversarial disruptions may become a threat to, including demand and supply prediction, grid operation, and disruption detectors. There emerged a notable gap in the amount of literature written about evasion disruptions compared to that written about poisoning disruptions, despite the fact that both are serious threats. Given this gap, our future work will be focused on understanding poisoning implications in power systems.  Specifically, we will focus on red teaming state-of-the-art data-driven methods applied to the power grid optimization. Additionally, our work will include multi-agent systems, as create more disruption surfaces, and they do not follow the traditional argument of secure centralized training. The grid is also trending towards multi-agent systems, and use of data-driven methods and renewables is increasing on the grid, so this work will become increasingly important. Data-driven optimization methods are becoming necessary at a time when the grid is facing increasing variability. It is critical to understand and address the vulnerabilities of these methods. This will make the grid more stable and give grid operators the reassurance they need to adopt these optimization methods.

\vspace{-.2cm}
\begin{credits}
\subsubsection{\ackname} This work is funded under AFOSR grant \#FA9550-24-1-0099. The views expressed are those of the authors and do not necessarily reflect the official policy or position of the Department of the Air Force, the Department of Defense, or the U.S. government.

\vspace{-.3cm}
\subsubsection{\discintname}
The authors have no competing interests to declare. 
\end{credits}

\vspace{-.2cm}
%
%
%

\end{document}